\RequirePackage{fix-cm}
\documentclass[smallextended]{svjour3}       
\smartqed  
\usepackage{graphicx}
\usepackage{mathtools}
\graphicspath{{images/}}
\usepackage[misc]{ifsym}
\begin{document}

\title{Automated detection of vulnerable plaque in intravascular ultrasound images
}


\author{Tae Joon Jun\textsuperscript{1}		\and
        Soo-Jin Kang\textsuperscript{2}		\and
        June-Goo Lee\textsuperscript{3}		\and
        Jihoon Kweon\textsuperscript{2}		\and
        Wonjun Na\textsuperscript{2}		\and
        Daeyoun Kang\textsuperscript{1}		\and
        Dohyeun Kim\textsuperscript{1}		\and
        Daeyoung Kim\textsuperscript{1}		\and
        Young-Hak Kim\textsuperscript{2}
}

\authorrunning{Tae Joon Jun et al.} 

\institute{Tae Joon Jun \\ 
           taejoon89@kaist.ac.kr
           \\\\
           \textsuperscript{1} School of Computing, Korea Advanced Institute of Science and Technology, Daejeon, Republic of Korea\\
           \textsuperscript{2} Division of Cardiology, University of Ulsan College of Medicine, Asan Medical Center, Seoul, Republic of Korea\\
           \textsuperscript{3} Asan Institute for Life Sciences, Asan Medical Center,
    Seoul, Republic of Korea
}

\date{Received: date / Accepted: date}

\maketitle

\begin{abstract}
Acute Coronary Syndrome (ACS) is a syndrome caused by a decrease in blood flow in the coronary arteries. The ACS is usually related to coronary thrombosis and is primarily caused by plaque rupture followed by plaque erosion and calcified nodule. Thin-cap fibroatheroma (TCFA) is known to be the most similar lesion morphologically to a plaque rupture. In this paper, we propose methods to classify TCFA using various machine learning classifiers including Feed-forward Neural Network (FNN), K-Nearest Neighbor (KNN), Random Forest (RF) and Convolutional Neural Network (CNN) to figure out a classifier that shows optimal TCFA classification accuracy. In addition, we suggest pixel range based feature extraction method to extract the ratio of pixels in the different region of interests to reflect the physician's TCFA discrimination criteria. A total of 12,325 IVUS images were labeled with corresponding OCT images to train and evaluate the classifiers. We achieved 0.884, 0.890, 0.878 and 0.933 Area Under the ROC Curve (AUC) in the order of using FNN, KNN, RF and CNN classifier. As a result, the CNN classifier performed best and the top 10 features of the feature-based classifiers (FNN, KNN, RF) were found to be similar to the physician's TCFA diagnostic criteria.
\end{abstract}

\section{Introduction}
\label{sec:1}

Acute coronary syndromes (ACS) is usually related to coronary thrombosis and is mainly caused by plaque rupture (55\%-60\%) followed by plaque erosion (30\%-35\%) and calcified nodule (2\%-7\%) \cite{virmani2006pathology}. Patients with ACS are more likely to undergo unstable angina, acute myocardial infarction, and sudden coronary death \cite{virmani2006pathology}. The most similar lesion morphologically to a plaque rupture, that is the most common type of vulnerable plaque, is known to be Thin-cap fibroatheroma (TCFA) which has a necrotic core and an underlying fibrous cap less than 65$\mu$m infiltrated by plenty of macrophages \cite{virmani2006pathology,kolodgie2001thin}. In order to observe this vulnerable plaque in the coronary arteries, Intravascular ultrasound (IVUS) and Optical Coherence Tomography (OCT) are generally used. IVUS provides a tomographic assessment of lumen area and conditions of the vessel wall. Besides that, IVUS additionally provides plaque size, distribution, and composition \cite{nissan2001intravascular}. Nevertheless, since the axial and lateral resolutions of IUVS are over 150$\mu$m it is hard to identify thin fibrous cap directly. Therefore, physicians mainly use OCT to identify TCFA lesion since the spatial resolution of OCT is less than 16$\mu$m which can visualize lipid-rich plaque and necrotic core \cite{jang2005vivo}. However, the length of images that can be observed at a single pull-back with OCT is limited to 35mm or less, and it is difficult to confirm the outline of the blood vessel since the blood vessel wall is usually not visible.

In this paper, we present a method for classifying TCFA, which is known to be the most common form of vulnerable plaque, using several machine learning algorithms. For feature-based classification, we used Feed-forward Neural Network (FNN), K Nearest Neighbor (KNN) classifiers and Random Forest (RF). In addition, we implemented a optimized classifier based on Convolutional Neural Networks (CNN) that uses IVUS images as direct input. The classification of TCFA is based on labels obtained by comparing 12,325 IVUS images from 100 patients with OCT images of the same frame. IVUS and OCT images were obtained from patients with either stable or unstable angina who underwent both IVUS and OCT procedures. After simultaneous IVUS and OCT images registration, the lumen and external elastic membrane (EEM) are segmented from the IVUS image to distinguish the areas of interest (ROI). For feature-based classifiers, we extracted 105 different features from each IVUS image, including the ratio of the intravascular plaque and the ratio of 10 pixels in four different plaque areas. Among the extracted features, N features selected through Fisher's Test are used as inputs of feature-based classifiers. For CNN classifier, 512 x 512 size greyscale IVUS images are randomly rotated for image augmentation, and these images are used as inputs to the optimized 20-depth CNN classifier. As a result, among the feature-based classifiers, the KNN classifier showed the best performance with 0.890 area under the curve (AUC), followed by AUC of 0.884 and 0.878 in the order of FNN and RF. The best results were obtained with a CNN classifier of 0.933 AUC, with 86.65\% Specificity and 83.08\% Sensitivity. One interesting result is that the top 10 features obtained using Fisher's exact test include the necrotic core ratio near the lumen and the ratio of the vascular plaque, which is also a reference to the physician's TCFA discrimination criteria. This paper is an improved and expanded version of the author's previous paper \cite{jun2017thincap}. The previous paper showed the possibility of classifying TCFA using the FNN classifier in OCT-guided IVUS images. In this paper, we improved the TCFA classification performance with optimized CNN classifier and evaluated a performance comparison with other machine learning algorithms.

Several previous studies related to the identification of TCFA have been reported. Jun proposed feed-forward neural network classifier to classify TCFA with IVUS and OCT images in \cite{jun2017thincap}. Prospective prediction of future development of TCFA with virtual histology IVUS (VH-IVUS) using support vector machine (SVM) was proposed by Zhang in \cite{zhang2015prospective}. Jang and Sawada reported in vivo characterization of TCFA with VH-IVUS and OCT in \cite{jang2005vivo,rodriguez2005vivo,sawada2008feasibility}. Association of VH-IVUS and major adverse cardiac events (MACE) on an individual plaque or whole patient analysis is presented by Calvert in \cite{calvert2011association}. Garcia-Garcia reported characteristics of coronary atherosclerosis including TCFA with IVUS and VH-IVUS in \cite{garcia2010imaging}. Recently, Inaba investigated the detection of pathologically defined TCFA using IVUS combined with Near-infrared spectroscopy (NIRS) in \cite{inaba2017intravascular}.

The paper is structured as follows. Section 2 contains specific methodologies for feature-based and CNN based TCFA classification. Section 3 contains the experimental setup and evaluation results for classification. Finally, Section 4 contains conclusions of this paper.

\section{Methodology}
\label{sec:2}
\begin{figure*}[!tbh]
	\centering
	\includegraphics[width=\textwidth]{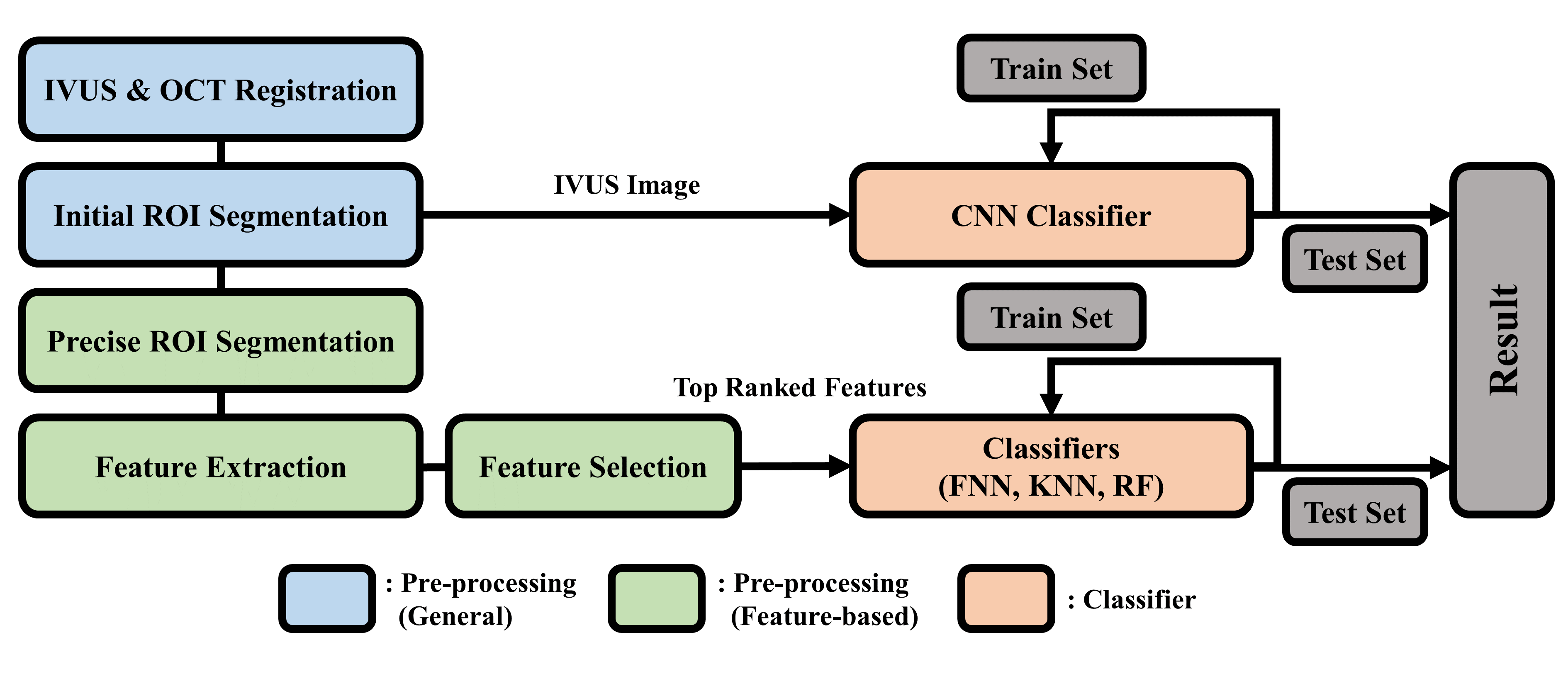}
	\caption{Overall procedures processed in TCFA classification}
	\label{Fig1}
\end{figure*}

Classifiers for TCFA classification include common pre-processing procedures, which are simultaneous IVUS and OCT images registration and ROI segmentation that distinguishes between lumen and EEM. A feature-based classifier is a classifier that uses manually extracted features which are extracted through precise ROI segmentation, feature extraction, and feature selection processes. In this paper, we used FNN, KNN, and RF as feature-based classifiers. The CNN classifier directly uses the pre-processed IVUS images as input and includes an image augmentation process to overcome the overfitting and lack of data in the training process. Even though the CNN classifier performs best, the reason for comparing performance with feature-based classifiers is that the feature-map of the CNN classifier to determine the TCFA differs from the criteria used by the physician to determine the TCFA. On the other hand, the manually extracted features used in the feature-based classifier are similar to the physician's TCFA discrimination criteria. For this reason, we classified TCFA using two methods: CNN classifier and feature-based classifiers. Fig. \ref{Fig1} illustrates the overall process of TCFA classification from pre-processing to classifier results.

\subsection{Data Pre-processing}
\label{sec:2.1}
\subsubsection{Simultaneous IVUS and OCT images registration}
\label{sec:2.1.1}
For classification of TCFA, labeled IVUS images are required, and these labeled images are obtained by an expert comparing the same IVUS and OCT image frames. Labeling IVUS image from equivalent OCT frame is called IVUS \& OCT co-registration. The above procedure is called simultaneous IVUS and OCT image registration. IVUS and OCT images of the same frame are obtained from patients who underwent both IVUS and OCT procedures with stable or unstable angina with lesions with angiographic 30 \% to 80 \% of diameter stenosis.

Within the target segment, all OCT sections at 0.2 mm intervals were registered with comparable IVUS frames (approximately 120th IVUS frame) using anatomical landmarks such as vessel shape, lateral branches, calcium, and perivascular structures, and distances from the ostium. Based on OCT images, the thinnest fibrous cap thickness of TCFA is less than 65$\mu$m and lipidic tissue angle is greater than 90 degrees. Each IVUS image is labeled with TCFA or normal based on the above criteria. Fig. \ref{Fig2} shows the overall process of simultaneous registration of IVUS and OCT images.

\subsubsection{Initial ROI Segmentation}
\label{sec:2.1.2}
\begin{figure}[!tbh]
	\centering
	\includegraphics[width=0.8\textwidth]{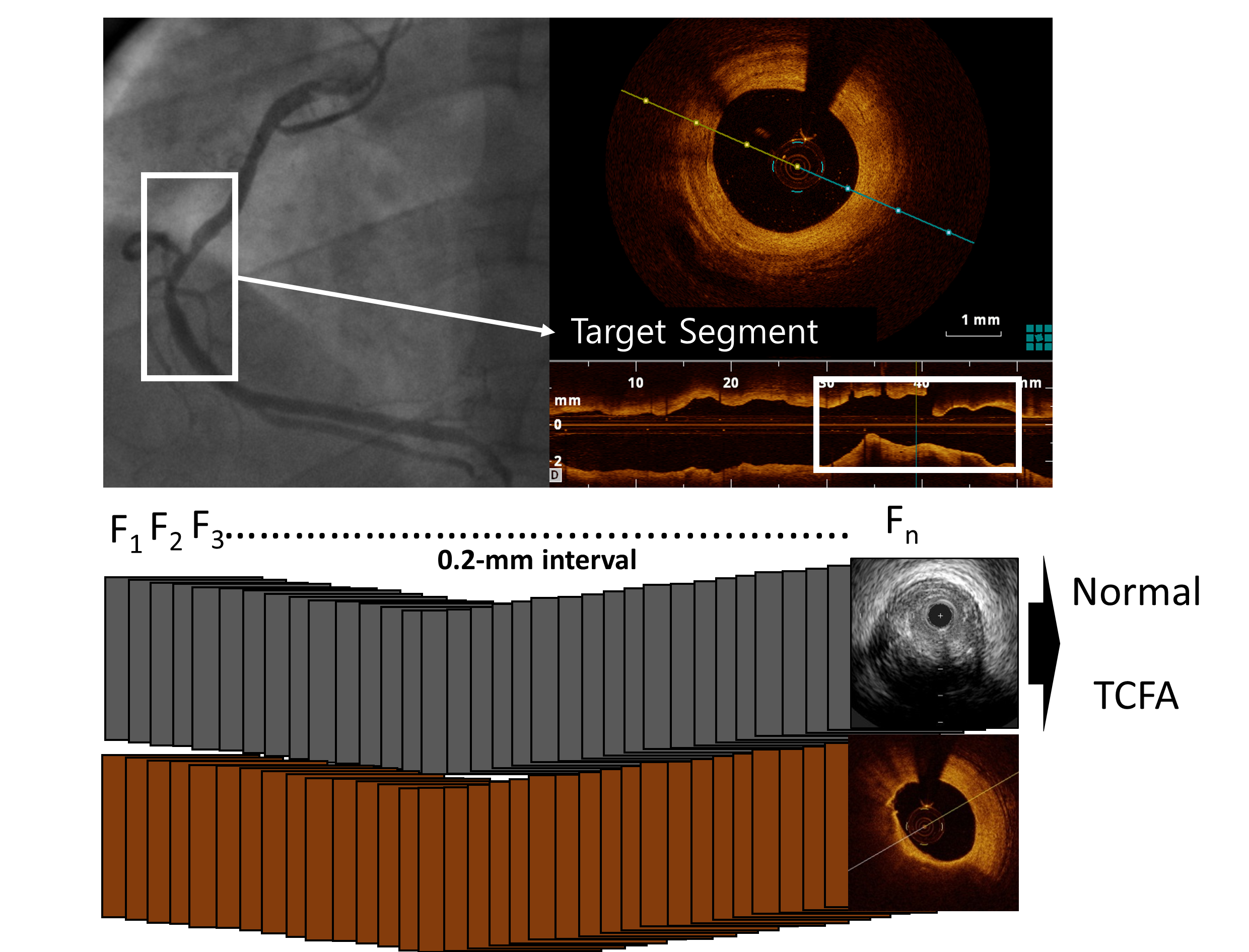}
	\caption{IVUS and OCT images registration \cite{jun2017thincap}}
	\label{Fig2}
\end{figure}

From the left figure of the Fig. \ref{Fig2}, initial ROI segmentation generates the mask image by dividing the original IVUS image into Adventitia, Lumen, and plaque. The three compartment divisions are conducted by the Medical Imaging Interaction Toolkit (MITK) \cite{wolf2005medical}. The feature-based classifier performs additional precise ROI segmentation process while the CNN classifier uses the image generated in the initial ROI segmentation process as an input.

\subsection{Feature based Classification}
\label{sec:2.2}
\subsubsection{Precise ROI Segmentation}
\label{sec:2.2.1}
\begin{figure}[!tbh]
	\centering
	\includegraphics[width=0.9\textwidth]{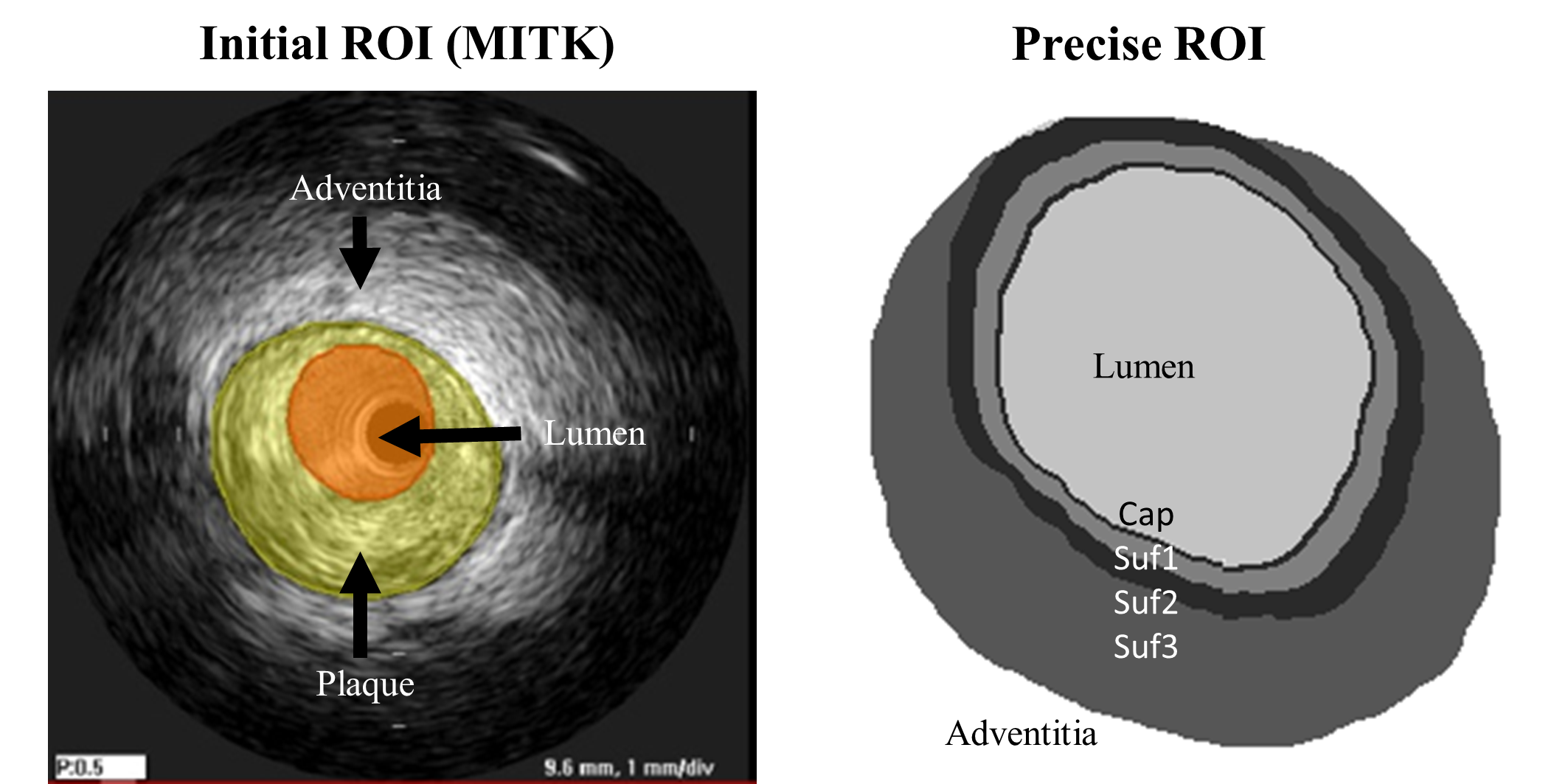}
	\caption{Initial (left) and precise (right) ROI segmentation}
	\label{Fig3}
\end{figure}

When determining TCFA, the importance of the distribution of necrotic core and lipid-rich core increases as it places more closer to the lumen. In particular, the IVUS image has an axial resolution of 150$\mu$m and a lateral resolution of 250$\mu$m, so TCFA cannot be directly detected since the criterion for TCFA is the thickness of fibrous cap of 65$\mu$m, which is smaller than IVUS resolution. Therefore, it is necessary to divide the initial ROI more precisely to compare the distribution of necrotic and lipid-rich cores. Moreover, it is known that distribution of necrotic and lipid core in the superficial region close to the lumen is important criteria when classifying TCFA. Therefore, the region near the lumen should be more precisely segmented. Through the Precise ROI segmentation process, we divide the plaque region into four different regions: Cap, Suf1, Suf2, and Suf3. Cap, the closest ROI to the lumen, contains an area of 2 pixels from the boundary between the lumen and the plaque. Next, Suf1, the second closest region to the lumen, contains the region between 2 pixels and 10 pixels, and similarly, Suf2 contains the region between 10 pixels and 20 pixels. Finally, Suf3 contains the entire region except for Cap, Suf1, and Suf2 in the plaque region. As a result, precise ROI segmentation extends the IVUS mask image, which was previously divided into three areas, into six areas. Fig. \ref{Fig3} shows 3 regions separated by initial ROI segmentation (left) and 6 extended regions separated by precise ROI segmentation.

\subsubsection{Feature Extraction}
\label{sec:2.2.2}
Obviously, the feature extraction procedure is the most important in determining the performance of a feature based classifier. Although it is not easy to recognize substance precisely in the IVUS image, the necrotic core and the lipid-rich core appear as relatively dark pixels while the calcified nodule appears as bright pixels Therefore, in this study, we tried to classify the TCFA based on the general pixel distribution information obtained from the IVUS image. In consideration of these characteristics, we extracted the pixel ratio of the four ROI regions as the feature of the classifier. We divide the pixel from 0 to 255 of the greyscale IVUS image into 26 areas with 10-pixel division unit. Therefore, the ratio of the 26-pixel areas belonging to each of the four ROI areas and the ratio of the plaques in the entire vessel is used as the input to the classifier. Consequently, 105(26*4+1) features were used total. Table. \ref{table_fe} shows a summarized description of 105 features for a 10-pixel division unit case. In Table. \ref{table_fe}, F1 represents the ratio of the plaque in the entire vessel described above, and F2 to F105 represents the ratio of pixel areas in each ROI region.
\begin{table}[!tbh]
	\caption{Feature description table}
    \centering
    \scalebox{1}{
	\begin{tabular}{lll}
    \hline\noalign{\smallskip}
		\textbf{    Feature    } &  \textbf{   Description   } & \textbf{   Region   }\\
        \noalign{\smallskip}\hline\noalign{\smallskip}
		F1 & Plaque / (Plaque + Lumen) & - \\
        F2 - F27 & Pixels{(0-10), (11-20), ... , (251-255)} & Cap \\
        F28 - F53 & Pixels{(0-10), (11-20), ... , (251-255)} & Suf1 \\
        F54 - F79 & Pixels{(0-10), (11-20), ... , (251-255)} & Suf2 \\
        F80 - F105 & Pixels{(0-10), (11-20), ... , (251-255)} & Suf3 \\
        \noalign{\smallskip}\hline
	\end{tabular}}
	\label{table_fe}
\end{table}

\subsubsection{Feature Selection}
\label{sec:2.2.3}
In general, if the number of features used as inputs increases beyond a certain limit, the performance of the classifier drops. This phenomenon is called a curse of dimensionality. This phenomenon is common to all feature-based classifiers used in this paper. For example, it is known that when neural networks are used for pattern recognition, the addition of irrelevant features beyond a certain point can degrade classification performance \cite{bishop1995neural}. Commonly used test methods for feature selection include Fisher's exact test and Chi-square test \cite{fisher1992statistical}, \cite{plackett1983karl}. However, it is known that chi-square test is better among the two test methods when the total number of samples is greater than 1,000. \cite{mcdonald2009handbook}. We, therefore, used the chi-square test to select the most relevant N number of features out of a total of 105 features and increase the size of the N from a minimum of 1 to a maximum of 105 to find an N with optimal classification performance. We also summarized the top 10 most relevant features that were found to be similar to the characteristics of coronary artery considered in the physician's TCFA diagnosis in Section 3.

\subsubsection{Feed-forward Neural Network Classifier}
\label{sec:2.2.4}
The first feature-based classifier used for TCFA classification is an FNN classifier, also known as multi-layer perceptron (MLP). we have improved the FNN classifier to demonstrate the faster convergence of TCFA classification over previous work \cite{jun2017thincap}. Our FNN classifier has five hidden layers, each of which consists of 50, 100, 200, 80, and 40 neurons. The rectified linear units (ReLU) is used as the activation function with alpha value 0.0001. Instead of using an Adam optimizer \cite{kingma2014adam} which we used earlier, we adopted a RMSprop optimizer \cite{tieleman2012lecture}. Before selected features were used as input, Min-max normalization was applied as pre-processing from 0 to 1.
\begin{equation}\label{(minmax)}
	\overline{x} = \frac{x-min}{Max-min}
\end{equation}
In the training phase, the initial learning rate starts at 0.001 and the batch size is set at 100. In each epoch, the learning rate gradually decreases exponentially by 0.95. This optimized FNN classifier shows TCFA classification performance similar to that of the previously proposed FNN but converges to the optimal result more quickly. Table \ref{table_fnn} summarizes the optimized parameters of the FNN classifier for TCFA classification.
\begin{table}[!tbh]
	\caption{Parameters for the FNN classifier}
    \centering
    \scalebox{1}{
	\begin{tabular}{ll}
    \hline\noalign{\smallskip}
		\textbf{    Parameter    }  & \textbf{   Value  }\\
        \noalign{\smallskip}\hline\noalign{\smallskip}
		Hidden Layers & 50 x 100 x 200 x 80 x 40 \\
        Activation & ReLU \\
        Optimizer & RMSprop \\
        Learning Rate & 0.001 (decay 0.95 / epoch) \\
        Batch Size & 100 \\
        \noalign{\smallskip}\hline
	\end{tabular}}
	\label{table_fnn}
\end{table}

\subsubsection{K-Nearest Neighbor Classifier}
\label{sec:2.2.5}
The second feature-based classifier used for TCFA classification is a KNN classifier. In case of using KNN classifier, the input consists of K-nearest training instances in the feature space, and the input is classified by the majority of its neighbors. As in the case of using the FNN classifier, Min-max normalization was applied to input features obtained through feature selection. The distance between the input sample and the training sample is calculated by using the Euclidean distance metric.
\begin{equation}\label{(euclidean)}
	d = \sqrt[]{\sum\limits_{i=1}^n(p_{i} - q_{i})^2}
\end{equation}
Another parameter to be considered in calculating the distance is the weight function. The weight function determines the weight given to the calculated distances for each neighbor, usually using the uniform weight function and the distance weight function. The uniform weight function assigns the same weight to all neighbors. On the other hand, the distance weight function assigns an inverse value of the calculated distance so that nearby neighbors can have a greater impact on classifying objects. A comparison of the two weight functions will also be discussed in Section 3 in conjunction with distance metrics. The last important parameter to be determined in the KNN classifier is K value. With the larger value of K, the impact caused by noise on categorizing entities becomes smaller. As a side effect, the boundaries needed to be classified into the class becomes ambiguous and classification results follow the class ratios of the training samples. Therefore, the method we used is to compare the performance of the TCFA classification through the validation set while increasing the K value to an odd number from 1 to 9 and apply the K value that has the best performance to the test set. The reason for testing odd values is to avoid ties when voting by majority vote. Table \ref{table_knn} summarizes the optimized parameters of the KNN classifier for TCFA classification.
\begin{table}[!tbh]\small
	\caption{Parameters for the KNN classifier}
    \centering
    \scalebox{1}{
	\begin{tabular}{ll}
    \hline\noalign{\smallskip}
		\textbf{    Parameter    }  & \textbf{   Value  }\\
        \noalign{\smallskip}\hline\noalign{\smallskip}
		Distance Metric & Euclidean \\
        Weight Function & Distance, Uniform \\
        K value & 1, 3, 5, 7, 9 \\
        \noalign{\smallskip}\hline
	\end{tabular}}
	\label{table_knn}
\end{table}

\subsubsection{Random Forest Classifier}
\label{sec:2.2.6}
The last feature-based classifier used for TCFA classification is a RF classifier which was first introduced by Breiman \cite{breiman2001random}. The RF classifier generates a number of decision trees and votes on the classes classified by each tree to determine the final class. It is a typical machine learning algorithm of Bootstrap Aggregation, also known as Bagging, which is a method of making N different prediction models randomly through N sampling and allowing individual prediction models to vote. As with the two classifiers above, Min-max normalization was applied to input features obtained through feature selection. The first parameter we considered when optimizing the RF classifier is the number of decision trees. If the number of decision trees is too small, the performance of the RF classifier drops, but if the number is too large, the training time increases. Therefore, we compared the performance of RF classifiers for 10, 50, and 100 different decision tree numbers. Another parameter to consider is the weight associated with the class. Since the number of normal data is greater than the number of TCFA, the class must be balanced through weighting to achieve fine performance in the RF classifier. Therefore, we adjust the weight inversely proportional to the class frequency of the input data. In case of the maximum depth of the RF classifier, the decision tree is expanded until all leaves have a pure value. Table \ref{table_rb} summarizes the optimized parameters of the RF classifier for TCFA classification.
\begin{table}[!tbh]\small
	\caption{Parameters for the RF classifier}
    \centering
    \scalebox{1}{
	\begin{tabular}{ll}
    \hline\noalign{\smallskip}
		\textbf{    Parameter    }  & \textbf{   Value  }\\
        \noalign{\smallskip}\hline\noalign{\smallskip}
		Number of Trees & 10, 50, 100 \\
        Class Weight & Inversely proportional to the frequency \\
        Max Depth & Expand until all leaves are pure \\
        \noalign{\smallskip}\hline
	\end{tabular}}
	\label{table_rb}
\end{table}

\subsection{Convolutional Neural Network classifier}
\label{sec:2.3}
\begin{figure}[!tbh]
	\centering
	\includegraphics[width=\textwidth]{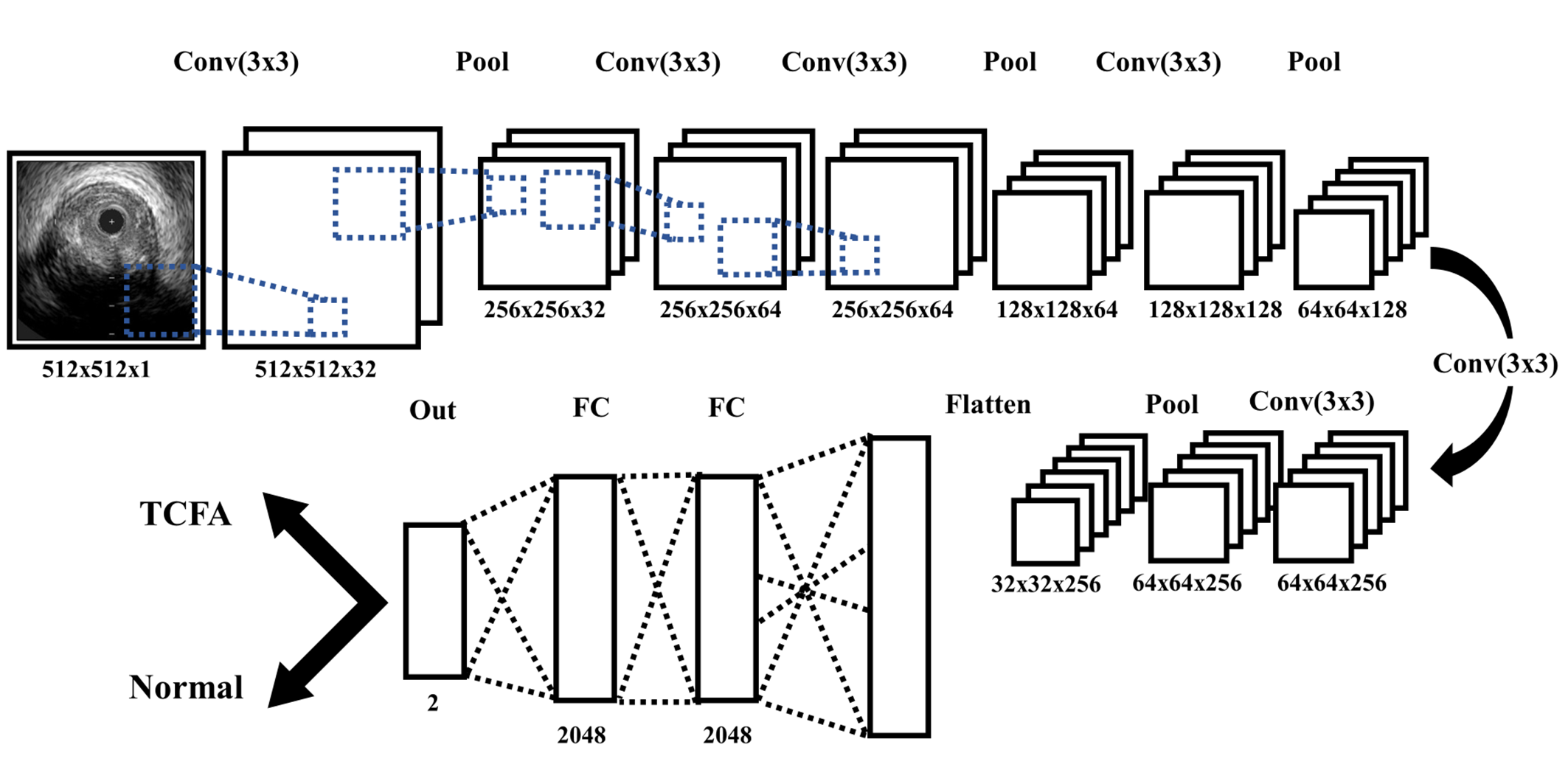}
	\caption{Proposed CNN classifier architecture}
	\label{Fig4}
\end{figure}

In this paper, we optimized CNN as the TCFA classifier. CNN was firstly introduced by LeCun et al\cite{lecun1989backpropagation} in 1989 and generalized by Simard et al \cite{simard2003best} in 2003. Nowadays, CNN is a state-of-art machine learning model in the field of pattern recognition, especially in image classification. There are several successful CNN models result from the ImageNet Large Scale Visual Recognition Challenge (ILSVRC)\cite{russakovsky2015imagenet} which is a competition for detecting and classifying objects from the given image set. Representative CNN models include AlexNet\cite{krizhevsky2012imagenet}, VGGNet\cite{simonyan2014very}, GoogLeNet\cite{szegedy2015going} and ResNet\cite{he2016deep}. The word `convolution' is originally used at filtering operation in the field of signal processing. Similarly, CNN extracts the feature map by iteratively processing convolution layers and pooling layers to the entire image. Different from the multi-layer neural network, also known as the fully connected neural network, CNN extracts the relation between the pixels that are spatially correlated by processing non-linear kernels to the image which dramatically reduces the number of free parameters and effectively handles an overfitting problem. With these multiple kernels and sub-sampling processes, CNN preserves the meaningful features from the local area of the image while reducing the size of the input image. In addition, since the multiple convolution layers are processed to the single image, eventually the CNN extracts a global feature with the chunk of local features. After bypassing multiple convolution layers and pooling layers, global feature map is used as an input layer for the fully connected layers. As a result, fully connected layers outputs two different probabilities of normal and TCFA classes and selects the maximum probability class as a classification result. Fig.\ref{Fig4} shows the detailed architecture of proposed CNN model.

The following procedures are important optimization techniques that we considered while constructing the proposed CNN model.

\subsubsection{Data augmentation}
\label{sec:2.3.1}
Data augmentation is one of the main advantages when using an image as an input data. When CNN is used as the classifier, data augmentation is highly recommended since enlarging the training data can effectively reduce overfitting and balance the distribution between the classes. Especially in the field of medical engineering, the most of the input data is normal (negative) while the number of diseases (positive) data is relatively smaller. In this case, the classifier that uses gradient descent method to reduce the loss function tends to optimize the model to fit the majority of data which is labeled as normal. Thus, the classifier may result in high specificity with low sensitivity. Therefore, balancing the train set with data augmentation can achieve both high specificity and sensitivity. In this work, we augmented TCFA images with randomly rotating the image by a multiple of 30 degrees. Unlike general image classification problem, we only augmented TCFA labeled images since the number of normal images is about 10 times larger than the number of TCFA images. By only augmenting the TCFA image, we can balance the normal and TCFA ratios of the train set of the classifier. In addition, we augmented the images by random rotation since the IVUS transducer rotates inside the vessel.

\subsubsection{Activation function}
\label{sec:2.3.2}
The role of activation function is to define output values from the kernel weights in CNN. In modern CNN model, non-linear activations including rectified linear unit (ReLU), leaky rectified linear unit (LReLU)\cite{maas2013rectifier}, and exponential linear unit (ELU)\cite{clevert2015fast} are widely used. While ReLU is the most popular activation function in CNN, LReLU and ELU impose slight negative values since converting entire negative values to zero in ReLU means that typical nodes may not participate anymore in further learning. Among LReLU and ELU, we chose ELU because it showed slightly better performance in the experiment. Therefore, we adopted ELU in our CNN model. ReLU, LReLU, and ELU are shown in following:

\begin{equation}\label{(3)}
	ReLU(x) = max(0,x)
\end{equation}

\begin{equation}\label{(4)}
	LReLU(x) = max(0,x) + \alpha min(0,x)
\end{equation}

\begin{equation}\label{(5)}
	\begin{gathered}
		ELU(x) = \bigg\{
		\begin{aligned}
			x\hspace{2cm}if\hspace{0.2cm}x\geq0\\
			\gamma(exp(x)-1)\hspace{1cm}if\hspace{0.2cm}x<0
		\end{aligned}
	\end{gathered}
\end{equation}

Where $\alpha$ is the leakage coefficient and $\gamma$ is the hyper-parameter to be tuned.

\subsubsection{Pooling Layer}
\label{sec:2.3.3}
In the neuroscience field, neurons tend to accept the strongest signal and ignore the others. Similarly, the CNN adopts this concept by using pooling layer. Pooling layer sub-samples spatially nearby pixels in the local feature maps. Generally, average pooling and max pooling are popularly used. However, when an average pooling is used in deep CNN, it reduces the impact of the strongest signal by averaging output values from the non-linear activation function. Therefore, max pooling is more widely used in modern deep CNN model. In addition, there is a stochastic pooling\cite{zeiler2013stochastic} method which maintains the advantage of max pooling method while reducing an overfitting effect. However, since our CNN model applied several efficient procedures to prevent an overfitting problem, we used max pooling as a pooling layer.

\subsubsection{Regularization}
\label{sec:2.3.4}
Regularization is a group of methods to prevent the overfitting problem. Traditional regularization methods are L1 and L2 regularization. Recently, dropout\cite{srivastava2014dropout} and batch normalization\cite{ioffe2015batch} are proposed to reduce the overfitting. Although, the original purpose of batch normalization is to reduce the internal covariate shift, preventing an overfitting is also known as the subsidiary effect of batch normalization. Therefore, we applied batch normalization at every convolution layer. Dropout is processed in the fully-connected layer with 0.5 probability of dropping the layer node. When the dropout is applied, it stops the node with given probability, resulting in prevention of nodes to co-adopt each other. Consequently, dropout leads model to reduce the overfitting.

\subsubsection{Cost and optimizer function}
\label{sec:2.3.5}
The cost function is minimized by using optimizer function. Generally, a cross-entropy cost function is used as a cost function.

\begin{equation}\label{(6)}
	C = - \frac{1}{n} \sum[y\ln a + (1-y)\ln (1-a)]
\end{equation}

Where \textit{n} is the number of training data (or the batch size), \textit{y} is an expected value, and \textit{a} is an actual value from the output layer.

Gradient descent based optimizer function with learning rate are used to minimize the cost function. There are several well-known optimizer functions such as RMSprop\cite{tieleman2012lecture}, Adam\cite{kingma2014adam}, Adagrad\cite{duchi2011adaptive}, and Adadelta\cite{zeiler2012adadelta}. In our CNN model, we adopted RMSprop optimizer function with 0.0001 starting learning rate by exponentially decaying the learning rate every 1,000 decay steps with 0.95 decay rate. Therefore, the learning rate at given global step can be computed as:

\begin{equation}\label{(7)}
	LR = LR_0 * 0.95^{\lfloor(GlobalStep/1,000)\rfloor}
\end{equation}

\subsubsection{Optimized CNN classifier architecture}
\label{sec:2.3.6}
Considering the above procedures, we designed a CNN classifier for TCFA classification. The main structure of the classifier resembles an ensemble of multiple VGGNets while optimizes several features to reduce the overfitting and to achieve higher classification accuracy. Table \ref{table_cnn} describes detailed parameters that we tuned in proposed CNN model.

\begin{table}[!tbh]
	\caption{Parameters for the CNN classifier}
    \centering
    \scalebox{1}{
	\begin{tabular}{ll}
    \hline\noalign{\smallskip}
		\textbf{    Parameter    }  & \textbf{   Value  }\\
        \noalign{\smallskip}\hline\noalign{\smallskip}
        Augmentation & random rotation by a multiple of 30 \\
		Kernel size & 3 x 3 (padding="same") \\
        Activation & ELU \\
        Optimizer & RMSprop \\
        Cost Function & Cross-entropy \\
        Regularization & Batch Normalization, Dropout \\
        Learning Rate & 0.001 (decay 0.95 / 1,000 steps) \\
        Batch Size & 32 \\
        \noalign{\smallskip}\hline
	\end{tabular}}
	\label{table_cnn}
\end{table}

\section{Experiments and Results}
\label{sec:3}
\subsection{Experimental Setup}
\label{sec:3.1}
We evaluated 100 patients who required clinical coronary angiography or percutaneous coronary intervention to participate in the study. All patients participating in the study provided written informed consent and the institutional review board of Asan Medical Center approved the study. OCT images for TCFA labeling were acquired with a C7XR system and LightLab Imaging's DragonFly catheter using a non-occlusion technique at a pullback rate of 20mm/s. Among the OCT images obtained by the above method, images with severe signal attenuation which are difficult to grasp the shape of the plaque were excluded from the study. The IVUS images used for TCFA classification were obtained with a pullback speed of 0.5mm/s using a motorized transducer from the Boston Scientific/SCIMED corporation which rotates in 40 MHz frequency with a 3.2-F imaging sheath. Within the IVUS images, only frames with a maximum plaque thickness of 0.5mm or more were included in the study. 

Through the above process, a total of 12,325 IVUS images that were labeled by matching OCT images were obtained. We randomly split 12,325 IVUS images into a training set and a test set at a ratio of 80 to 20 while keeping the TCFA expression rates of the two sets similar. We used Python language to develop classifiers. The scikit-learn library \cite{pedregosa2011scikit} was used mainly when implementing KNN and RF classifiers. For the FNN classifier, the Keras and scikit-learn libraries are mostly used. In case of developing the CNN classifier, higher computational power is required than the previous three classifiers since the raw 512 x 512 greyscale IVUS image is used as an input. Therefore we used GPU version TensorFlow library \cite{abadi2016tensorflow} with two NVIDIA Titan X GPU which has 3,584 CUDA cores and 12GB of GPU memory.

The performance evaluation of the TCFA classification was based on the following three metrics: Area Under the Curve (AUC), Specificity, and Sensitivity. AUC refers to the size of the area under the Receiver Operating Characteristic (ROC) curve, and the closer the AUC value is to 1, the better the performance of the classifier. If the AUC value is 0.5, it means that the classifier can not classify normal and TCFA at all since it is the same as the result when randomly classified as normal or TCFA. Table. \ref{table_guide} presents the guideline rules for interpreting the AUC value suggested by Hosmer and Lemeshow \cite{hosmer2013applied}. 

\begin{table}[!tbh]\small
	\caption{AUC interpretation guidelines}
    \centering
    \scalebox{1}{
	\begin{tabular}{ll}
    \hline\noalign{\smallskip}
		\textbf{AUC} & \textbf{Guidelines} \\
        \noalign{\smallskip}\hline\noalign{\smallskip}
        0.5 - 0.6 & No discrimination\\
        0.6 - 0.7 & Poor discrimination\\
        0.7 - 0.8 & Acceptable discrimination\\
        0.8 - 0.9 & Good discrimination\\
        0.9 - 1 & Excellent discrimination\\
        \noalign{\smallskip}\hline
	\end{tabular}}
	\label{table_guide}
\end{table}

Specificity, also known as the true negative rate, measures the percentage of negatives that are correctly identified as normal. Sensitivity, also known as the true positive rate or recall, measures the percentage of positives that are correctly identified as TCFA. Specificity is the ratio of negative test results that are correctly classified as normal. Sensitivity is the probability of positive test results that are correctly identified as TCFA. These two metrics are defined with following four terminologies from a confusion matrix:

\begin{itemize}
\item True Positive(TP): The number of patients correctly identified as TCFA
\item False Positive(FP): The number of patients incorrectly identified as TCFA
\item True Negative(TN): The number of patients correctly identified as normal
\item False Negative(FN): The number of patients incorrectly identified as normal
\end{itemize}

\begin{equation}\label{(1)}
	Specificity(Sp) = \frac{TN}{FP + TN} \times 100 (\%)
\end{equation}

\begin{equation}\label{(2)}
	Sensitivity(Se) = \frac{TP}{TP + FN} \times 100 (\%)
\end{equation}

\subsection{Feature-based classifiers evaluation}
\label{sec:3.2}
In evaluating the TCFA classification, specificity and sensitivity are inversely proportional to each other. Generally, the optimal cut-off value in the ROC curve is obtained when the sum of specificity and sensitivity is the maximum. Therefore, we applied above criteria when assessing the classification performance of the classifiers. In case of evaluating the feature-based classifiers, a different number of input features result in different AUC value and ROC curve. As a result, we can obtain an AUC graph with the number of input features on the x-axis. Consequently, the final specificity and sensitivity are calculated with the optimal cut-off value from the ROC curve when the AUC value is the maximum. Table \ref{result_table_summ} shows the summarized evaluation results of the three feature-based classifiers and the result of Jun et al \cite{jun2017thincap} where N refers to the number of features when the AUC value is the maximum.

\begin{table}[!tbh]\small
	\caption{Evaluation result of Feature-based classifiers}
    \centering
    \scalebox{1}{
	\begin{tabular}{lllll}
    \hline\noalign{\smallskip}
		\textbf{} & \textbf{N} & \textbf{AUC} & \textbf{Sp(\%)} & \textbf{Se(\%)}\\
        \noalign{\smallskip}\hline\noalign{\smallskip}
		{    Jun et al \cite{jun2017thincap}   } & 73 & 0.868 & 78.31 & 79.02\\
        {    FNN   } & 82 & 0.884 & 75.18 & 87.80\\
		{    KNN    } & 77 & 0.890 & 79.38 & 86.83\\
        {    RF   } & 102 & 0.878 & 76.41 & 88.29\\
        \noalign{\smallskip}\hline
	\end{tabular}}
	\label{result_table_summ}
\end{table}

\subsubsection{FNN classifier evaluation}
\label{sec:3.2.1}
\begin{figure}[!tbh]
	\centering
	\includegraphics[width=\textwidth]{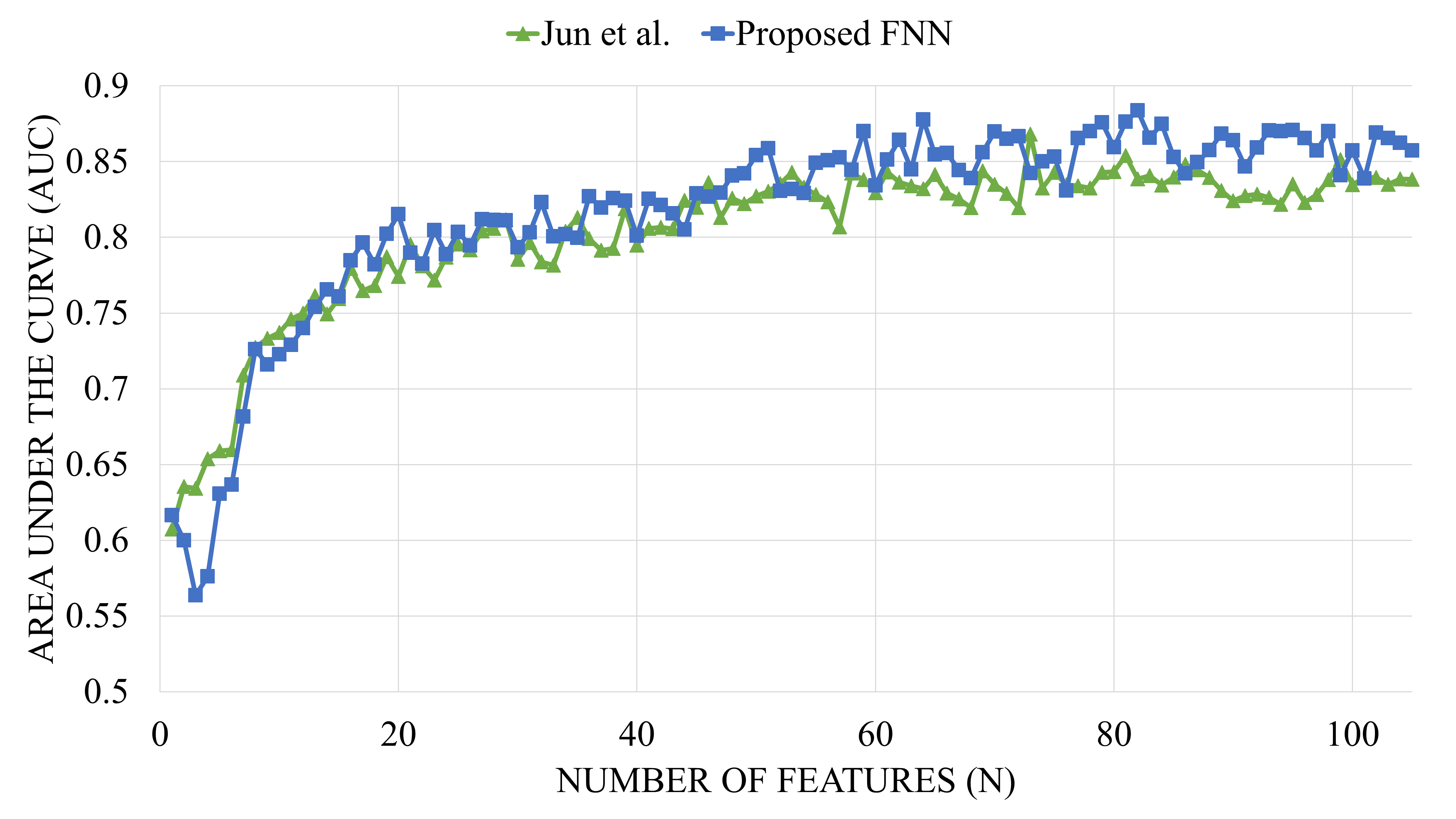}
	\caption{AUC graph from the proposed FNN classifier}
	\label{Fig5}
\end{figure}

Fig. \ref{Fig5} describes the AUC graph of proposed FNN classifier. From the Fig. \ref{Fig5}, the maximum AUC value of 0.884 is obtained when the number of features is 82. When all 105 features are used as inputs, the AUC value is 0.857 which is 3.1\% lower than the best AUC value. The specificity and sensitivity of optimal cut-off value are 75.18\% and 87.80\%. The proposed FNN classifier achieved a higher AUC of 1.84\% compared to the results of Jun et al \cite{jun2017thincap}.

\subsubsection{KNN classifier evaluation}
\label{sec:3.2.2}
\begin{figure}[!tbh]
	\centering
	\includegraphics[width=\textwidth]{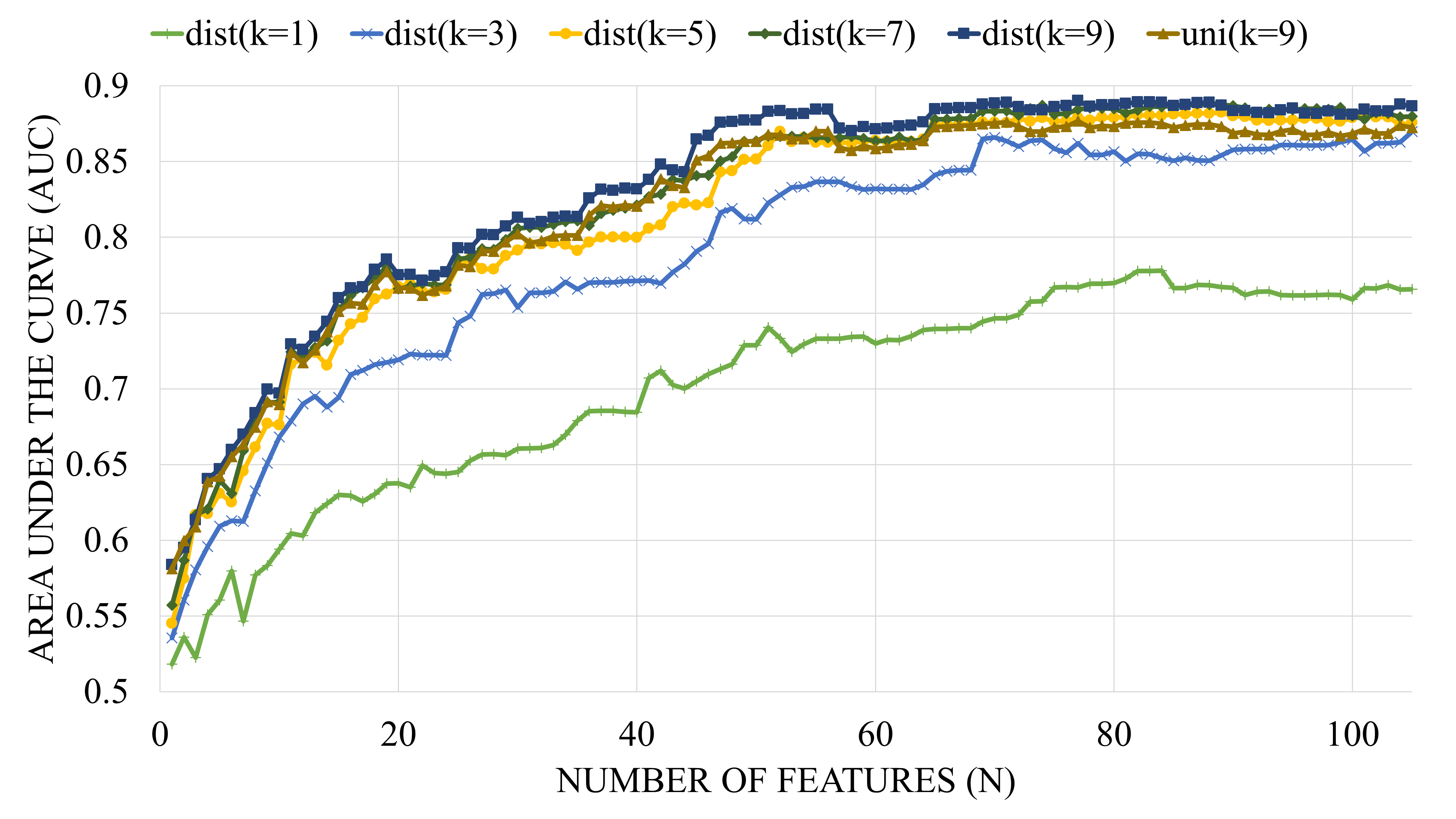}
	\caption{AUC graph from the proposed KNN classifier}
	\label{Fig6}
\end{figure}

Fig. \ref{Fig6} shows the AUC graph of proposed KNN classifier. From the Fig. \ref{Fig6}, the maximum AUC value of 0.890 is obtained when the number of features is 77 and the K value of 9 with distance weight function (\textbf{dist(k=9)}). In case of using the uniform weight function with the K value of 9 (\textbf{uni(k=9)}), the maximum AUC value is 0.877 which is 2.3\% lower than the best AUC value. The specificity and sensitivity of optimal cut-off value are 79.38\% and 86.83\%.

\subsubsection{RF classifier evaluation}
\label{sec:3.2.3}
\begin{figure}[!tbh]
	\centering
	\includegraphics[width=\textwidth]{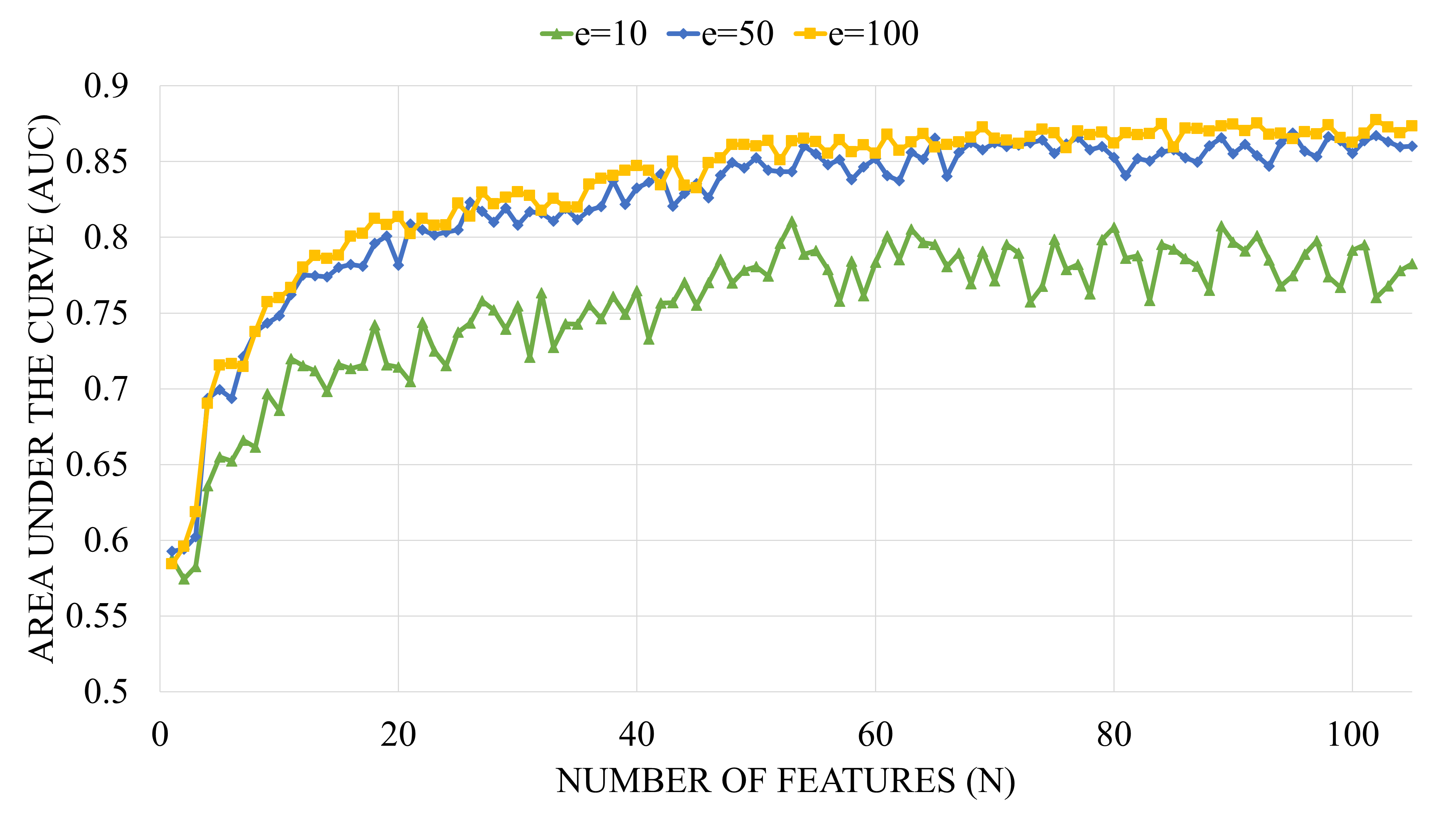}
	\caption{AUC graph from the proposed RF classifier}
	\label{Fig7}
\end{figure}

Fig. \ref{Fig7} presents the AUC graph of proposed RF classifier. From the Fig. \ref{Fig7}, the maximum AUC value of 0.878 is obtained when the number of features is 102 with the 100 decision trees (\textbf{e=100}). In case of using 50 decision trees (\textbf{e=50}), the maximum AUC value is 0.869 while the value is 0.811 when 10 decision trees are used (\textbf{e=10}). Therefore, it seems that more than 50 decision trees should be used to obtain a reasonable result. The specificity and sensitivity of optimal cut-off value are 76.41\% and 88.29\%.

\subsection{CNN classifier evaluation}
\label{sec:3.3}
\begin{figure}[!tbh]
	\centering
	\includegraphics[width=\textwidth]{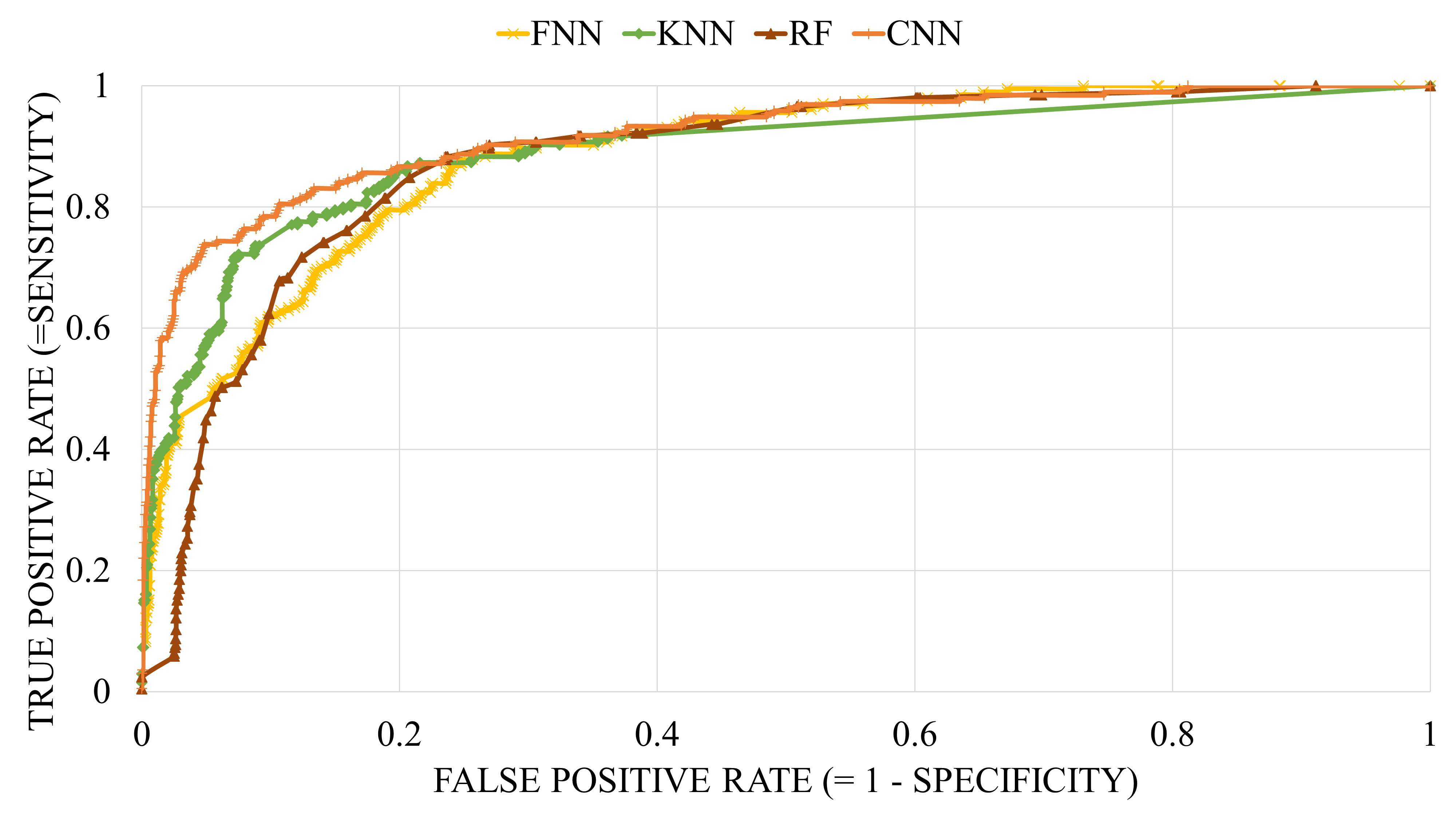}
	\caption{ROC Curve graph from the results of proposed classifiers}
	\label{Fig8}
\end{figure}

To evaluate the TCFA classification with CNN classifier, we set 10\% of the training set as a validation set. This validation set evaluates the loss and accuracy of the model every epoch. If the loss of the model does not improve for 3 epochs, we stop the training and evaluate the model with the test set. After training the model 16 epochs, validation loss improved from 0.17. Therefore, we evaluated the model with test set based on the weights saved at 16th epochs. As a result, we obtained the maximum AUC value of 0.933 and the specificity and sensitivity of optimal cut-off value were 86.65\% and 83.08\%. Table \ref{result_table_cnn} presents the evaluation results of the proposed classifiers, and Fig. \ref{Fig8} represents the ROC curve of the CNN evaluation result including the results from three feature-based classifiers. 

\begin{table}[!tbh]\small
	\caption{Evaluation result of proposed classifiers}
    \centering
    \scalebox{1}{
	\begin{tabular}{lllll}
    \hline\noalign{\smallskip}
		\textbf{} & \textbf{AUC} & \textbf{Sp(\%)} & \textbf{Se(\%)}\\
        \noalign{\smallskip}\hline\noalign{\smallskip}
		{    CNN    } & 0.933 & 86.65 & 83.08\\
        {    KNN    } & 0.890 & 79.38 & 86.83\\
        {    FNN    } & 0.884 & 75.18 & 87.80\\
        {    RF    } & 0.878 & 76.41 & 88.29\\
        {    Jun et al \cite{jun2017thincap}    } & 0.868 & 78.31 & 79.02\\
        \noalign{\smallskip}\hline
	\end{tabular}}
	\label{result_table_cnn}
\end{table}

\subsection{Feature Ranking evaluation}
\label{sec:3.4}
Chi-square test was performed on the training set and a score list of 105 features was obtained according to the p-value. Table. \ref{table_rank} shows the top 10 ranked features which have a correlation with TCFA class. First, we can see that there is a high correlation with TCFA in that intra-vascular plaque ratio (F1) is the second highest. This is fairly obvious since the more plaques exist in an artery, the higher chance of getting TCFA the artery will have. Secondly, it can be seen that the ratio of the relatively dark pixels (0 - 30) near the lumen (Cap) and TCFA class is highly correlated (F2, F3, F4). This is related to the necrotic core which appears relatively dark in the IVUS image. Finally, the bright pixels (141 - 180) in the superficial plaque region (Suf1) also have some correlation with TCFA class. This seems to be related to the calcified region but requires further confirmation. In addition to the top 10 features, rest of features are found to have a correlation with the TCFA class of about 40\%.

\begin{table}[!tbh]\small
	\caption{Top 10 ranked features }
    \centering
    \scalebox{1}{
	\begin{tabular}{lllll}
    \hline\noalign{\smallskip}
		\textbf{    Rank    } & \textbf{    Feature    } & \textbf{   Ratio(\%)   } & \textbf{   Pixel range   } & \textbf{   Region   }\\
        \noalign{\smallskip}\hline\noalign{\smallskip}
		1 & F2 & 18.64 & 0 - 10 & Cap \\
		2 & F1 & 17.52 & - & - \\
        3 & F3 & 4.75 & 11 - 20 & Cap \\
        4 & F41 & 3.69 & 151 - 160 & Suf1 \\
        5 & F42 & 3.48 & 161 - 170 & Suf1 \\
        6 & F40 & 2.62 & 141 - 150 & Suf1 \\
        7 & F43 & 2.17 & 171 - 180 & Suf1 \\
        8 & F4 & 2.14 & 21 - 30 & Cap \\
        9 & F69 & 1.69 & 111 - 120 & Suf2 \\    
        10 & F15 & 1.61 & 151 - 160 & Cap \\
        etc & - & 41.69 & - & - \\
        \noalign{\smallskip}\hline
	\end{tabular}}
	\label{table_rank}
\end{table}

\section{Conclusion}
\label{sec:4}
In this paper, we propose a method to classify TCFA using several machine learning classifiers. Our proposed classifier includes feature-based classifiers and deep learning classifier. In order to classify TCFA with IVUS images, IVUS and OCT registration process and initial ROI segmentation were performed as a common pre-processing. In case of using feature-based classifiers, input features are required to be extracted from the raw IVUS images. Therefore, we suggest pixel range based feature extraction method to extract the ratio of pixels in the different region of interests. With the extracted features above, we selected the most correlated features to the TCFA class by using Chi-square test on training data set. These selected features are trained with feature-based classifiers that include feed-forward neural network, K-nearest neighbor, and random forest classifiers. By training the optimized classifier, we achieved AUC value of 0.884, 0.890 and 0.878 in order of FNN, KNN and RF classifiers. In case of using deep learning classifier, we optimized the convolutional neural network classifier which is a cutting-edge model in the field of image classification. By training the CNN classifier with an augmented IVUS image, we obtained AUC value of 0.933 which is improved than the results of feature-based classifiers. We also analyzed the top-ranked features from the Chi-square test to find the basis for classifier's TCFA detection. We believe that this research will help in the study of machine learning with IVUS images in vascular.

\begin{acknowledgements}
This research was supported by International Research \& Development Program of the National Research Foundation of Korea(NRF) funded by the Ministry of Science, ICT\&Future Planning of Korea(2016K1A3A7A03952054) and support of Asan Medical Center providing IVUS images and clinical advices for this research are gratefully acknowledged.
\end{acknowledgements}



\end{document}